\documentclass[10pt,twocolumn,letterpaper]{article}

\usepackage{cvpr}
\usepackage{times}
\usepackage{epsfig}
\usepackage{graphicx}
\usepackage{amsmath}
\usepackage{amssymb}
\usepackage[pagebackref=true,breaklinks=true,letterpaper=true,colorlinks,bookmarks=false]{hyperref}

\cvprfinalcopy 

\usepackage{amsmath,amsfonts,bm}

\def\eqref#1{equation~\ref{#1}}

\def\1{\bm{1}}

\DeclareMathAlphabet{\mathsfit}{\encodingdefault}{\sfdefault}{m}{sl}
\SetMathAlphabet{\mathsfit}{bold}{\encodingdefault}{\sfdefault}{bx}{n}

\DeclareMathOperator*{\argmax}{arg\,max}

\usepackage{hyperref}
\usepackage{url}

\usepackage{graphicx}
\usepackage{enumitem}
\usepackage[export]{adjustbox}
\usepackage{tabularx}
\usepackage{array}
\usepackage{booktabs}  
\usepackage{subcaption} 
\usepackage{multirow} 

\usepackage{hyperref}
\hypersetup{
    colorlinks=true,
    linkcolor=blue,
    filecolor=magenta,      
    urlcolor=cyan,
}

\newcommand*{\affmark}[1][*]{\textsuperscript{#1}}

\usepackage{xcolor}
\newcommand{\minisection}[1]{\vspace{0.04in}\noindent {\bf #1}\ \ }

\title{MineGAN: effective knowledge transfer from  GANs to target domains with few images}

\author{Yaxing Wang\affmark[1], Abel Gonzalez-Garcia\affmark[1], David Berga\affmark[1]\\ 
Luis Herranz\affmark[1], Fahad  Shahbaz Khan\affmark[2,3], Joost van de Weijer\affmark[1] \\
\affmark[1]~Computer Vision Center, Universitat Aut\`onoma de Barcelona, Spain\\
\affmark[2]~Inception Institute of Artificial Intelligence, UAE  \affmark[3]~CVL, Link\"oping University, Sweden\\
{\tt\small \{yaxing,agonzalez,dberga,lherranz,joost\}@cvc.uab.es, fahad.khan@liu.se}
}

\begin{document}

\maketitle

\begin{abstract}
One of the attractive characteristics of deep neural networks is their ability to transfer knowledge obtained in one domain to other related domains. As a result, high-quality networks can be trained in domains with relatively little training data. This property has been extensively studied for discriminative networks but has received significantly less attention for generative models.  
Given the often enormous effort required to train GANs, both computationally as well as in the dataset collection, the re-use of pretrained GANs is a desirable objective.
We propose a novel knowledge transfer method for generative models based on mining the knowledge that is most beneficial to a specific target domain, either from a single or multiple pretrained GANs. 
This is done using a miner network that identifies which part of the generative distribution of each pretrained GAN outputs samples closest to the target domain. 
Mining effectively steers GAN sampling towards suitable regions of the latent space, which facilitates the posterior finetuning and avoids pathologies of other methods such as mode collapse and lack of flexibility.
We perform experiments on several complex datasets using various GAN architectures (BigGAN, Progressive GAN) and show that the proposed method, called MineGAN, effectively transfers knowledge to domains with few target images, outperforming existing methods. 
In addition, MineGAN can successfully transfer knowledge from multiple pretrained GANs. Our code is available at: \url{https://github.com/yaxingwang/MineGAN}.
\end{abstract}

\vspace{-2mm}
\section{Introduction}
\vspace{-1mm}
Generative adversarial networks (GANs) can learn the complex underlying distribution of image collections~\cite{goodfellow2014generative}. They have been shown to generate high-quality realistic images~\cite{karras2017progressive,karras2018style,brock2018large} and are used in many applications including image manipulation~\cite{pix2pix2017,zhu2017unpaired},  style transfer~\cite{gatys2016image}, compression~\cite{tschannen2018deep}, and colorization~\cite{zhang2016colorful}. 
It is known that high-quality GANs require a significant amount of training data and time. For example, Progressive GANs~\cite{karras2017progressive} are trained on 30K images and are reported to require a month of training on one NVIDIA Tesla V100.
Being able to exploit these high-quality pretrained models,
not just to generate the distribution on which they are trained, but also to combine them with other models and adjust them to a target distribution is a desirable objective. For instance, it might be desirable to only generate women using a GAN trained to generate men and women alike. Alternatively, one may want to generate smiling people from two pretrained generative models, one for men and one for women. The focus of this paper is on performing these operations using only a small target set of images, and without access to the large datasets used to pretrain the models.

Transferring knowledge to domains with limited data has been extensively studied for discriminative models~\cite{donahue2014decaf,oquab2014learning,pan2010survey, tzeng2015simultaneous}, enabling the re-use of high-quality networks.
However, knowledge transfer for generative models has received significantly less attention, possibly due to its great difficulty, especially when transferring to target domains with few images.
Only recently,Wang et al.~\cite{wang2018transferring} studied finetuning from a single pretrained generative model and showed that it is beneficial for domains with scarce data. However, Noguchi and Harada~\cite{noguchi2019image} observed that this technique leads to mode collapse. 
Instead, they proposed to reduce the number of trainable parameters, and only finetune the learnable parameters for the batch normalization (scale and shift) of the generator. 
Despite being less prone to overfitting, their approach severely limits the flexibility of the knowledge transfer.

In this paper, we address knowledge transfer by adapting a trained generative model for targeted image generation given a small sample of the target distribution.
We introduce the process of \emph{mining} of GANs. This is performed by a \emph{miner network} that transforms a multivariate normal distribution into a distribution on the input space of the pretrained GAN in such a way that the generated images resemble those of the target domain. The miner network has considerably fewer parameters than the pretrained GAN and is therefore less prone to overfitting.
The mining step predisposes the pretrained GAN to sample from a narrower region of the latent distribution that is closer to the target domain, which in turn eases the subsequent finetuning step by providing a cleaner training signal with lower variance (in contrast to sampling from the whole source latent space as in~\cite{wang2018transferring}).
Consequently, our method preserves the adaptation capabilities of finetuning while preventing overfitting. 

Importantly, our mining approach enables transferring from multiple pretrained GANs, which allows us to aggregate information from multiple sources simultaneously to generate samples akin to the target domain. We show that these networks can be trained by a selective backpropagation procedure.  Our main contributions are: 
\setlist{nolistsep}
\begin{itemize}[noitemsep]
       \item We introduce a novel miner network to steer the sampling of the latent distribution of a pretrained GAN to a target distribution determined by few images. 
        \item We propose the first method to transfer knowledge from multiple GANs to a single generative model.
        \item We outperform existing competitors on a variety of settings, including transferring knowledge from unconditional, conditional, and multiple GANs. 
\end{itemize}

\section{Related work}
\minisection{Generative adversarial networks.} GANs consists of two modules: generator and discriminator~\cite{goodfellow2014generative}. 
The generator aims to generate images to fool the discriminator, while the discriminator aims to distinguish generated from real images. Training GANs was initially difficult, due to mode collapse and training instability.  
Several methods focus on addressing these problems~\cite{gulrajani2017improved,salimans2016improved,mao2017least,arjovsky2017wasserstein,miyato2018spectral}, while another major line of research aims to improve the architectures to generate higher quality images~\cite{radford2015unsupervised,denton2015deep,karras2017progressive,karras2019style,brock2018large}. 
For example, Progressive GAN~\cite{karras2017progressive} generates better images by synthesizing them progressively from low to high-resolution. 
Finally, BigGAN~\cite{brock2018large} successfully performs conditional high-realistic generation from ImageNet~\cite{deng2009imagenet}.

\minisection{Transfer learning for GANs.} While knowledge transfer has been widely studied for discriminative models in computer vision~\cite{donahue2014decaf,pan2010survey, oquab2014learning,tzeng2015simultaneous}, only a few works have explored transferring knowledge for generative models~\cite{noguchi2019image,wang2018transferring}. 
Wang \etal~\cite{wang2018transferring} investigated finetuning of pretrained GANs, leading to improved performance for target domains with limited samples. This method, however, suffers from mode collapse and overfitting, as it updates all parameters of the generator to adapt to the target domain. Recently, Noguchi and Harada~\cite{noguchi2019image} proposed to only update the batch normalization parameters. 
Although less susceptible to mode collapse, this approach significantly reduces the adaptation flexibility of the model since changing only the parameters of the batch normalization permits for style changes but is not expected to function when shape needs to be changed. 
They also replaced the GAN loss with a mean square error loss. As a result, their model only learns the relationship between latent vectors and sparse training samples, requiring the input noise distribution to be truncated during inference to generate realistic samples. 
The proposed MineGAN does not suffer from this drawback, as it learns how to automatically adapt the input distribution. 
In addition, we are the first to consider transferring knowledge from multiple GANs to a single target domain. 

\minisection{Iterative image generation.} Nguyen et al.~\cite{nguyen2016synthesizing} have investigated training networks to generate images that maximize the activation of neurons in a pretrained classification network. 
In a follow-up approach~\cite{nguyen2017plug} that improves the diversity of the generated images, they use this technique to generate images of a particular class from a pretrained classifier network. 
In principle, these works do not aim at transferring knowledge to a new domain, and can instead only be applied to generate a distribution that is exactly described by one of the class labels of the pretrained classifier network. Another major difference is that the generation at inference time of each image is an iterative process of successive backpropagation updates until convergence, whereas our method is feedforward during inference.

\begin{figure*}[t]
    \centering
    \includegraphics[width=\textwidth]{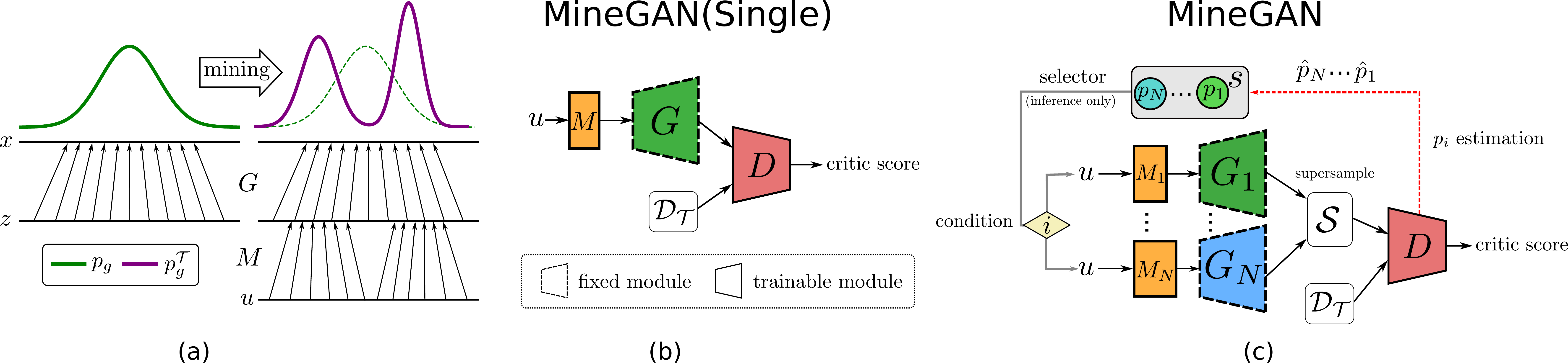}\vspace{-3mm}
    \caption{\small (a) Intuition behind our approach for a simple case. Mining shifts the prior input distribution towards the most promising regions with respect to given target data $\mathcal{D}_\mathcal{T}$. In practice, the input distribution is much more complex. (b) Architecture implementing the proposed mining operation on a single GAN. Miner $M$ identifies the relevant regions of the prior distribution so that generated samples are close to the target data $\mathcal{D}_\mathcal{T}$. Note that during the first stage, when training the miner, the generator remains fixed. In a second stage, we finetune the miner, generator and discriminator together. (c) Training setup for multiple generators. Miners $M_1$,...,$M_N$ identify subregions of the pretrained generators while selector $\mathcal{S}$ learns the sampling frequencies of the various generators. 
    }\vspace{-5mm}    \label{fig:models}
\end{figure*}

\section{Mining operations on GANs}
Assume we have access to one or more pretrained GANs and wish to use their knowledge to train a new GAN for a target domain with few images. 
For clarity's sake, we first introduce mining from a single GAN in Section~\ref{sec:singleGAN}, but our method is general for an arbitrary number of pretrained GANs, as explained in Section~\ref{sec:multipleGANs}. Then, we show how the miners can be used to train new GANs (Section~\ref{sec:mineGAN}).

\subsection{GAN formulation}
Let $p_{data}(x)$ be a probability distribution over real data $x$ determined by a set of real images $\mathcal{D}$, and let $p_z(z)$ be a prior distribution over an input noise variable $z$.
The generator $G$ is trained to synthesize images given $z \sim p_z(z)$ as input, inducing a generative distribution $p_g(x)$ that should approximate the real data distribution $p_{data}(x)$.
This is achieved through an adversarial game~\cite{goodfellow2014generative}, in which a discriminator $D$ aims to distinguish between real images and images generated by $G$, while the generator tries to generate images that fool $D$.
In this paper, we follow WGAN-GP~\cite{gulrajani2017improved}, which provides better convergence properties by using the Wasserstein loss~\cite{arjovsky2017wasserstein} and a gradient penalty term (omitted from our formulation for simplicity). The discriminator (or critic) and generator losses are defined as follows:
\begin{equation}
\mathcal{L}_{D}= \mathbb{E}_{z \sim p_z(z)}[D(G(z))]
- \mathbb{E}_{x\sim p_{data}(x)}[D(x)] ,
\end{equation}
\vspace{-4mm}
\begin{equation}
\mathcal{L}_{G} = - \mathbb{E}_{z \sim p_z(z)}[D(G(z))].
\end{equation}

We also consider families of pretrained generators $\{G_i\}$.
Each $G_i$ has the ability to synthesize images given input noise $z\sim p_z^i(z)$.
For simplicity and without loss of generality, we assume the prior distributions are Gaussian, i.e.\ $p_z^i(z) = \mathcal{N}(z|\bm{\mu_i}, \bm{\Sigma_i})$.
Each generator $G_i(z)$ induces a learned generative distribution $p_g^i(x)$, which approximates the corresponding real data distribution $p_{data}^i(x)$ over real data $x$ given by the set of source domain images $\mathcal{D}_i$.

\subsection{Mining from a single GAN}
\label{sec:singleGAN}

We want to approximate a target real data distribution $p_{data}^\mathcal{T}(x)$ induced by a set of real images $\mathcal{D}_\mathcal{T}$, given a critic $D$ and a generator $G$, which have been trained to approximate a source data distribution $p_{data}(x)$ via the generative distribution $p_g(x)$.
The mining operation learns a new generative distribution $p_g^\mathcal{T}(x)$ by finding those regions in $p_g(x)$ that better approximate the target data distribution $p_{data}^\mathcal{T}(x)$ while keeping $G$ fixed.
In order to find such regions, mining actually finds a new prior distribution $p_z^\mathcal{T}(z)$ such that samples $G(z)$ with $z \sim p_z^\mathcal{T}(z)$ are similar to samples from $p_{data}^\mathcal{T}(x)$ (see Fig.~\ref{fig:models}a).
For this purpose, we propose a new GAN component called \emph{miner}, implemented by a small multilayer perceptron $M$. 
Its goal is to transform the original input noise variable $u \sim p_z(u)$ to follow a new, more suitable prior that identifies the regions in $p_g(x)$ that most closely align with the target distribution.

Our full method acts in two stages. 
The first stage steers the latent space of the fixed generator $G$ to suitable areas for the target distribution. 
We refer to the first stage as \emph{MineGAN (w/o FT)} and present the proposed mining architecture in Fig.~\ref{fig:models}b.
The second stage updates the weights of the generator via finetuning ($G$ is no longer fixed). 
\emph{MineGAN} refers to our full method including finetuning.

Miner $M$ acts as an interface between the input noise variable and the generator, which remains fixed during training.
To generate an image, we first sample $u \sim p_z(u)$, transform it with $M$ and then input the transformed variable to the generator, i.e.\ $G(M(u))$. 
We train the model adversarially: the critic $D$ aims to distinguish between fake images output by the generator $G(M(u))$ and real images $x$ from the target data distribution $p^\mathcal{T}_{data}(x)$.
We implement this with the following modification on the WGAN-GP loss:
\vspace{-2mm}
\begin{equation}
\mathcal{L}_D^{M}= \mathbb{E}_{u \sim p_z(u)}[D(G(M(u)))]
- \mathbb{E}_{x\sim p^\mathcal{T}_{data}(x)}[D(x)] ,\label{eq_min1}
\end{equation}
\vspace{-4mm}
\begin{equation}
\mathcal{L}_{G}^{M} = - \mathbb{E}_{u \sim p_z(u)}[D(G(M(u)))].\label{eq_min2}
\end{equation}
The parameters of $G$ are kept unchanged but the gradients are backgropagated all the way to $M$ to learn its parameters.
This training strategy will gear the miner towards the most promising regions of the input space, i.e.\ those that generate images close to $\mathcal{D}_\mathcal{T}$.
Therefore, $M$ is effectively mining the relevant input regions of prior $p_z(u)$ and giving rise to a targeted prior $p_z^\mathcal{T}(z)$, which will focus on these regions while ignoring other ones that lead to samples far off the target distribution $p_{data}^\mathcal{T}(x)$.

We distinguish two types of targeted generation: on-manifold and off-manifold.
In the \emph{on-manifold} case, there is a significant overlap between the original distribution $p_{data}(x)$ and the target distribution $p_{data}^\mathcal{T}(x)$.
For example, $p_{data}(x)$ could be the distribution of human faces (both male and female) while $p_{data}^\mathcal{T}(x)$ includes female faces only. 
On the other hand, in \emph{off-manifold} generation, the overlap between the two distributions is negligible, e.g.\ $p_{data}^\mathcal{T}(x)$ contains cat faces.
The off-manifold task is evidently more challenging as the miner needs to find samples out of the original distribution (see Fig.~\ref{fig:target_women_children}).
Specifically, we can consider the images in $\mathcal{D}$ to lie on a high-dimensional image manifold that contains the support of the real data distribution $p_{data}(x)$~\cite
{arjovsky2017iclr}.
For a target distribution farther away from $p_{data}(x)$, its support will be more disjoint from the original distribution's support, and thus its samples might be off the manifold that contains $\mathcal{D}$.

\subsection{Mining from multiple GANs} 
\label{sec:multipleGANs}
In the general case, the mining operation is applied on multiple pretrained generators.
Given target data $\mathcal{D}_\mathcal{T}$, the task consists in mining relevant regions from the induced generative distributions learned by a family of $N$ generators $\{G_i\}$.
In this task, we do not have access to the original data used to train $\{G_i\}$ and can only use target data $\mathcal{D}_\mathcal{T}$.
Fig.~\ref{fig:models}c presents the architecture of our model, which extends the mining architecture for a single pretrained GAN by including multiple miners and an additional component called \emph{selector}.
In the following, we present this component and describe the training process in detail.

\minisection{Supersample.}
In traditional GAN training, a fake minibatch is composed of fake images $G(z)$ generated with different samples $z\sim p_z(z)$.
To construct fake minibatches for training a set of miners, we introduce the concept of \emph{supersample}.
A supersample $\mathcal{S}$ is a set of samples composed of exactly one sample per generator of the family, i.e.\ $\mathcal{S} = \{G_i(z) | z\sim p_z^i(z); i=1,...,N\}$.
Each minibatch contains $K$ supersamples, which amounts to a total of $K\times N$ fake images per minibatch. 

\minisection{Selector.}
The selector's task is choosing which pretrained model to use for generating  samples during inference.
For instance, imagine that $\mathcal{D}_1$ is a set of `kitchen' images and $\mathcal{D}_2$ are `bedroom' images, and let  $\mathcal{D}_\mathcal{T}$ be `white kitchens'.
The selector should prioritize sampling from $G_1$, as the learned generative distribution $p_g^1(x)$ will contain kitchen images and thus will naturally be closer to $p_{data}^\mathcal{T}(x)$, the target distribution of white kitchens.
Should $\mathcal{D}_\mathcal{T}$ comprise both white kitchens and dark bedrooms, sampling should be proportional to the distribution in the data. 

We model the selector as a random variable $s$ following a categorical distribution parametrized by
$p_1,...,p_N$ with $p_i>0$, $\sum p_i = 1$.
We estimate the parameters of this distribution as follows.
The quality of each sample $G_i(z)$ is evaluated by a single critic $D$ based on its critic value $D(G_i(z))$. Higher critic values indicate that the generated sample from $G_i$ is closer to the real distribution. 
For each supersample $\mathcal{S}$ in the minibatch, we record which generator obtains the maximum critic value, i.e.\ $\argmax_{i} D(G_i(z))$.
By accumulating over all $K$ supersamples and normalizing, we obtain an empirical probability value $\hat{p}_i$ that reflects how often generator $G_i$ obtained the maximum critic value among all generators for the current minibatch. We estimate each parameter $p_i$ as the empirical average $\hat{p}_i$ in the last 1000 minibatches.
Note that $p_i$ are learned during training and fixed during inference, where we apply a multinomial sampling function  to sample the index.

\minisection{Critic and miner training.} We now define the training behavior of the remaining learnable components, namely the critic $D$ and miners $\{M_i\}$, when
minibatches are composed of supersamples. 
The critic aims to distinguish real images from fake images. This is done by looking for artifacts in the fake images which distinguish them from the real ones. 
Another less  discussed but equally important task of the critic is to observe the frequency of occurrence of images: if some (potentially high-quality) image occurs more often among fake images than real ones, the critic will lower its score, and thereby motivate the generator to lower the frequency of occurrence of this image.  
Training the critic by backpropagating from all images in the supersample prevents it from assessing the frequency of occurrence of the generated images and leading to unsatisfactory results empirically. 
Therefore, the loss for multiple GAN mining is: 
\vspace{-2mm}
\begin{equation}
\begin{split}
\mathcal{L}_D^{M}= \mathbb{E}_{ \{ u^{i} \sim p_z^{i}(u) \}}[\underset{i}{\max} \{D(G_i(M_i(u^i)))\}]\\
- \mathbb{E}_{x\sim p^\mathcal{T}_{data}(x)}[D(x)] 
\end{split}
\end{equation}
\vspace{-4mm}
\begin{equation}
\mathcal{L}_{G}^{M} = - \mathbb{E}_{\{u^{i} \sim p_z^{i}(u)\}}[\underset{i}{\max}\{D(G_i(M_i(u^i)))\}].\label{eq_multiMGAN}
\end{equation}
As a result of the $\max$ operator we only backpropagate from the generated image that obtained the highest critic score. This allows the critic to assess the frequency of occurrence correctly. Using this strategy, the critic can perform both its tasks: boosting the quality of images as well as driving the miner to closely follow the distribution of the target set. 
Note that we initialize the single critic $D$ with the pretrained weights from one of the pretrained critics\footnote{We empirically found that starting from any pretrained critic leads to similar results~(see Suppl. Mat. (Sec.~F))}. 

\begin{figure}[t]
    \centering
    \includegraphics[width=0.9\columnwidth]{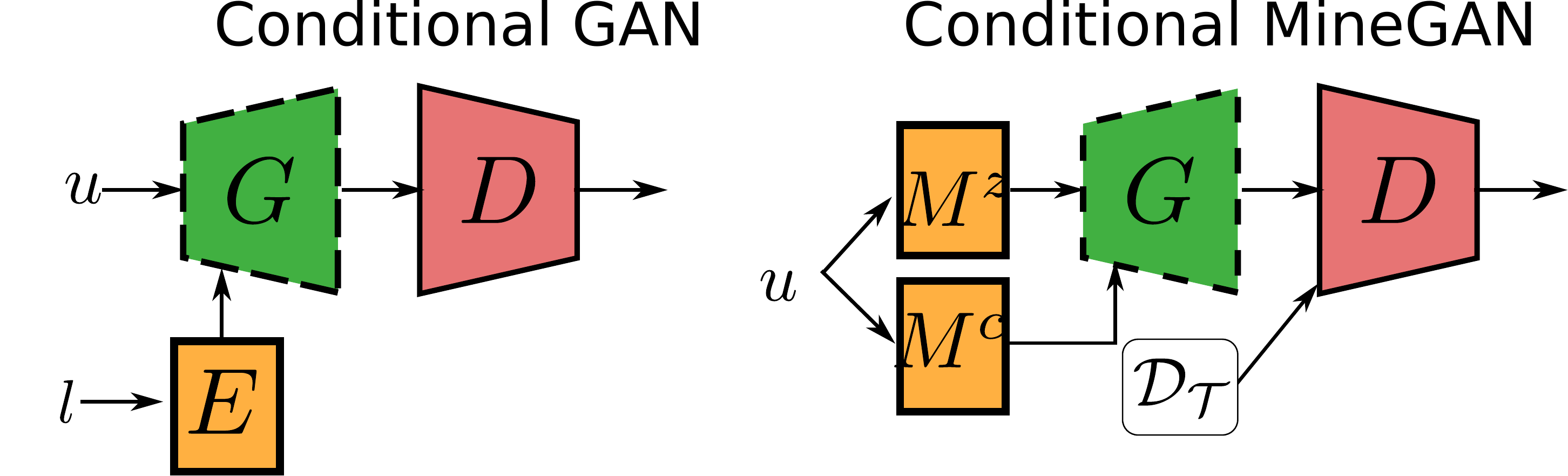}
    \caption{\small Application of mining in conditional setting (on BigGAN~\cite{brock2018large}). We apply an additional miner network to estimate the class embedding. $\mathcal{D}_\mathcal{T}$: target data, $E$: class embedding, $l$: label. }\vspace{-5mm}
    \label{fig:condition_models}
\end{figure}

\minisection{Conditional GANs.} So far, we have only used unconditional GANs.
However, conditional GANs (cGANs), which introduce an additional input variable to condition the generation to the class label, are used by the most successful approaches~\cite{brock2018large,zhang2018self}.
Here we extend our proposed MineGAN to cGANs that condition on the batch normalization layer~\cite{dumoulin2016learned,brock2018large}\footnote{See Suppl. Mat. (Sec.~D) for results on another type of conditioning.}, more concretely, BigGAN~\cite{brock2018large} (Fig.~\ref{fig:condition_models} (left)).
First, a label $l$ is mapped to an embedding vector by means of a class embedding $E$, and then this vector is mapped to layer-specific batch normalization parameters. The discriminator is further conditioned via label projection~\cite{miyato2018cgans}. 
Fig.~\ref{fig:condition_models} (right) shows how to mine BigGANs.
Alongside the standard miner $M^z$, we introduce a second miner network $M^c$, which maps from $u$ to the embedding space, resulting in a generator $G(M^c(u),M^z(u))$. The training is  equal to that of a single GAN and follows Eqs.~\ref{eq_min1} and~\ref{eq_min2}.

\subsection{Knowledge transfer with MineGAN}
\label{sec:mineGAN}
The underlying idea of mining is to predispose the pretrained model to the target distribution by reducing the divergence between source and target distributions.
The miner network contains relatively few parameters and is therefore less prone to overfitting, which is known to occur when directly finetuning the generator $G$~\cite{noguchi2019image}. 
We finalize the knowledge transfer to the new domain by finetuning both the miner $M$ and generator $G$ (by releasing its weights). 
The risk of overfitting is now diminished as the generative distribution is closer to the target, requiring thus a lower degree of parameter adaptation.
Moreover, the training is substantially more efficient than directly finetuning the pretrained GAN~\cite{wang2018transferring}, where synthesized images are not necessarily similar to the target samples. A mined pretrained model makes the sampling more effective,  leading to less noisy gradients and a cleaner training signal.

\section{Experiments}
\label{sec:experiments}
We first introduce the used evaluation measures and architectures.
Then, we evaluate our method for knowledge transfer from unconditional GANs, considering both single and multiple pretrained generators. 
Finally, we assess transfer learning from conditional GANs. Experiments focus on transferring knowledge to target domains with few images. 

\minisection{Evaluation measures.} We employ the widely used Fr\'echet Inception Distance (FID)~\cite{heusel2017gans} for evaluation. FID measures the similarity between two sets in the embedding space given by the features of a convolutional neural network. 
More specifically, it computes the differences between the estimated means and covariances assuming a multivariate normal distribution on the features.
FID measures both the quality and diversity of the generated images and has been shown to correlate well with human perception~\cite{heusel2017gans}. 
However, it suffers from instability on small datasets.
For this reason, we also employ Kernel Maximum Mean Discrepancy (KMMD) with a Gaussian kernel and Mean Variance (MV) for some experiments~\cite{noguchi2019image}.
Low KMMD values indicate high quality images, while high values of MV indicate more image diversity.

\minisection{Baselines.}
We compare our method with the following baselines.
\textit{TransferGAN~\cite{wang2018transferring}} 
directly updates both the generator and the discriminator for the target domain.
\textit{VAE~\cite{kingma2013auto}}
is a variational autoencoder trained following~\cite{noguchi2019image}, i.e.\ fully supervised by pairs of latent vectors and training images. 
\textit{BSA~\cite{noguchi2019image}} updates only the batch normalization parameters of the generator instead of all the parameters.
\textit{DGN-AM~\cite{nguyen2016synthesizing}} generates images that maximize the activation of neurons in a pretrained classification network.
\textit{PPGN~\cite{nguyen2017plug}} improves the diversity of DGN-AM  by of adding a prior to the latent code via denoising autoencoder.
Note that both of DGN-AM and PPGN require the target domain label, and thus we only include them in the conditional setting.

\minisection{Architectures.} We apply mining to  Progressive GAN~\cite{karras2017progressive}, SNGAN~\cite{miyato2018spectral}, and BigGAN~\cite{brock2018large}. 
The miner has two fully connected layers for MNIST and four layers for all other experiments.
More training details in Suppl. Mat. (Sec.~A).

\subsection{Knowledge transfer from unconditional GANs}
\vspace{-2mm}
\label{sec:uncoditional}

\begin{figure}[t]
    \centering
    \begin{tabular}{ c c } 
    \includegraphics[width=.15\textwidth,height=1.5cm]{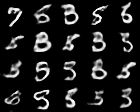}
    \includegraphics[width=.15\textwidth,height=1.5cm]{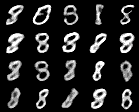}
    \includegraphics[width=.15\textwidth,height=1.5cm]{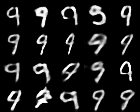}\\
    \end{tabular}
    \vspace{-3mm}
    \caption{\small Results for off-manifold generation of MineGAN(w/o FT). We generate 20 samples of digits `5', `8' or `9'.\vspace{-5mm}}
    \label{fig:target_MNIST}
\end{figure}

\minisection{MNIST dataset.}
We show results on MNIST to illustrate the functioning of the miner~\footnote{We add quantitative results on MNIST in Suppl. Mat.~(Sec.~D)}.
We use 1000 images of size $28 \times 28$ as target data. We test mining for off-manifold targeted image generation. 
In off-manifold targeted generation, $G$ is pre-trained to synthesize all MNIST digits except for the target one, e.g.\ $G$ generates 0-8 but not 9.  The results in Fig.~\ref{fig:target_MNIST} are after training only the miner, without an additional finetuning step.
Interestingly, the miner manages to steer the generator to output samples that resemble the target digits, by merging patterns from other digits in the source set.
For example, digit `9' frequently resembles a modified 4 while `8' heavily borrows from 0s and 3s. 
Some digits can be more challenging to generate, for example, `5' is generally more distinct from other digits and thus in more cases the resulting sample is confused with other digits such as `3'. In conclusion, even though target classes are not in the training set of the pretrained GAN, still similar examples might be found on the manifold of the generator.

\begin{figure*}[t]
    \centering
    \includegraphics[width=\textwidth]{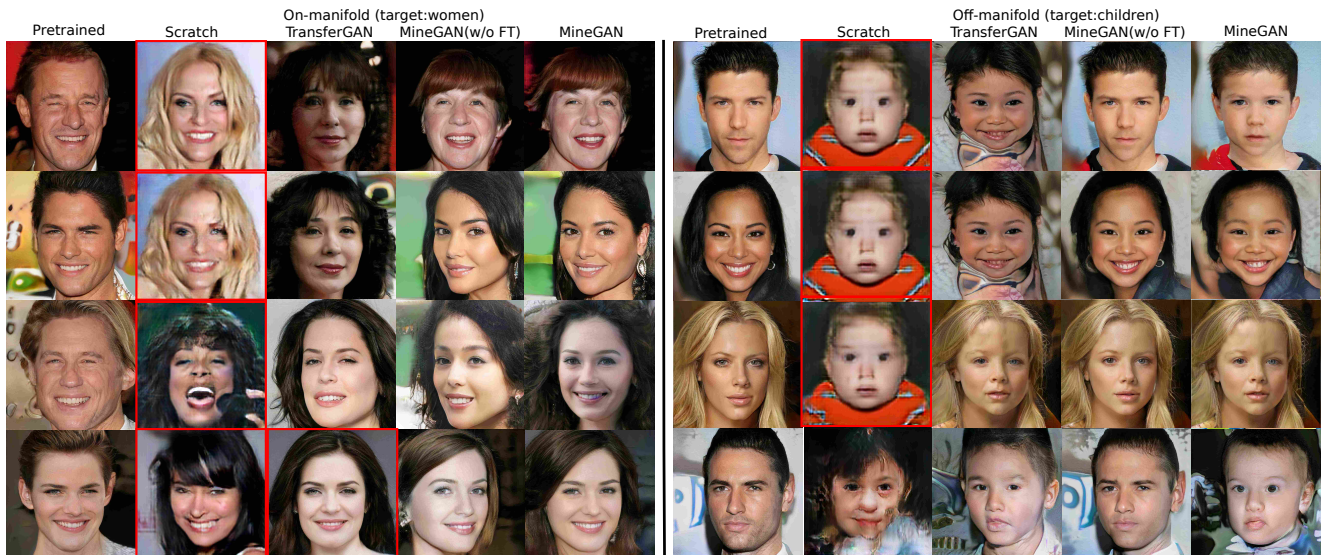}\vspace{-3mm}
    \caption{\small Results: (Left) On-manifold (CelebA$\rightarrow$FFHQ women),  (Right) Off-manifold (CelebA$\rightarrow$FFHQ children). Based on pretrained Progressive GAN. The images in red boxes suffer from overfitting. See Suppl. Mat. (Figs.~11-14 and Sec.~E) for more examples.}
    \label{fig:target_women_children}
\end{figure*}

\begin{figure*}[t]
    \centering
    \includegraphics[width=1\textwidth]{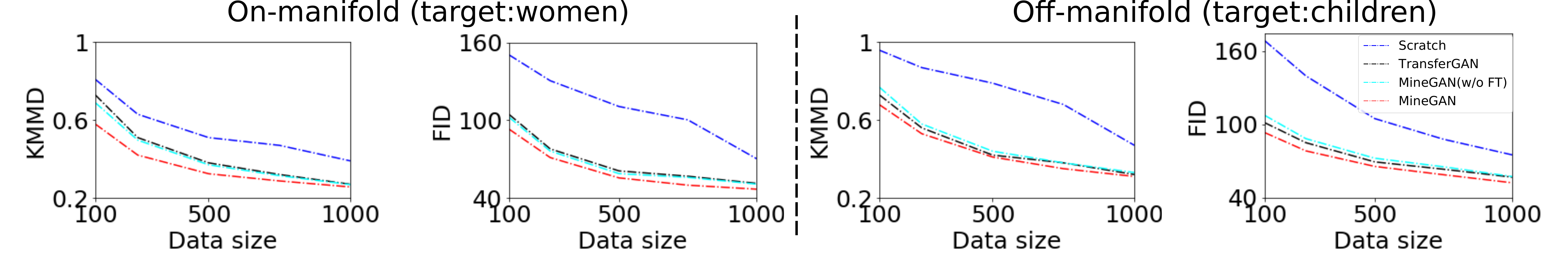}\vspace{-3mm}
    \caption{\small KMMD and FID on CelebA$\rightarrow$FFHQ women (left) and CelebA$\rightarrow$FFHQ children (right).}\vspace{-4mm}    \label{fig:target_women_children_fid}
\end{figure*}

\minisection{Single pretrained model.} We start by transferring knowledge from a Progressive GAN trained on \textit{CelebA~\cite{liu2015deep}}. 
We evaluate the performance on target datasets of varying size, using $1024 \times 1024$ images. 
We consider two target domains: on-manifold, \textit{FFHQ women}~\cite{karras2019style} and off-manifold, \textit{FFHQ children face}~\cite{karras2019style}. 
We refer as \emph{MineGAN} to our full method including finetuning, whereas \emph{MineGAN (w/o FT)} refers to only applying mining (fixed generator). We use training from \emph{Scratch}, and the \emph{TransferGAN} method of~\cite{wang2018transferring} as baselines. 
Figure~\ref{fig:target_women_children_fid} shows the performance in terms of FID and KMMD as a function of the number of images in the target domain. 
MineGAN outperforms all baselines. 
For the on-manifold experiment, MineGAN (w/o FT) outperforms the baselines, and results are further improved with additional finetuning. 
Interestingly, for the off-manifold experiment, MineGAN (w/o FT) obtains only slightly worse results than TransferGAN, showing that the miner alone already manages to generate images close to the target domain. 
Fig.~\ref{fig:target_women_children} shows images generated when the target data contains 100 training images. 
Training the model from scratch results in overfitting, a pathology also occasionally suffered by TransferGAN.
MineGAN, in contrast, generates high-quality images without overfitting and images are sharper, more diverse, and have more realistic fine details.

\begin{figure}[t]
\centering
	\def\arraystretch{0.5}
    \begin{adjustbox}{max width=\textwidth}
    
    \footnotesize{
    \begin{tabular}{m{0.5\textwidth}m{0.5\textwidth}}
    \includegraphics[width=0.5\textwidth]{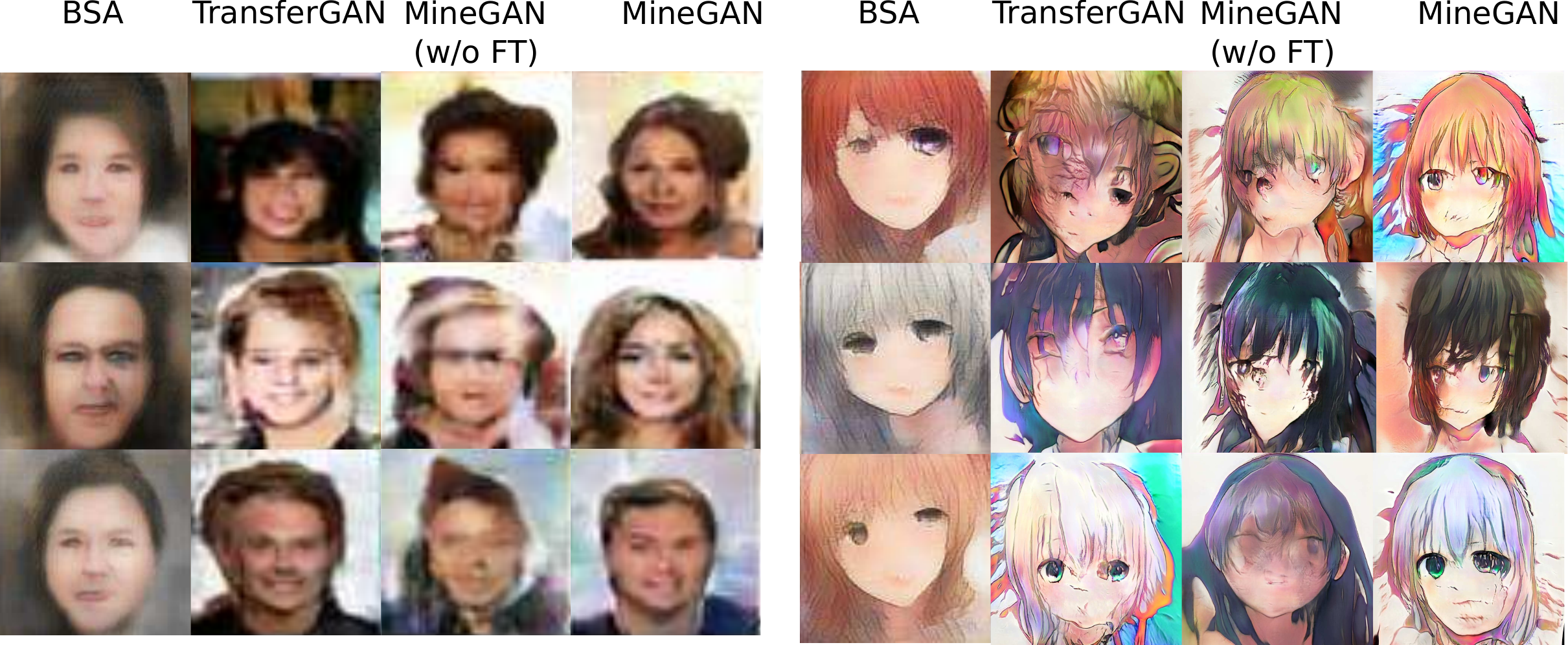}& \\         \resizebox{1\linewidth}{!}{
            \begin{tabular}{lcccc}
    \toprule
    \multirow{2}{*}{Method} & \multicolumn{2}{c}{FFHQ} & \multicolumn{2}{c}{Anime Face} \\
       & KMMD & MV    & KMMD & MV \\
     \midrule
    From scratch & 0.890 & -& 0.753 & -  \\
    TransferGAN~\cite{wang2018transferring} & 0.346 & 0.506 & 0.347 & 0.785 \\
    VAE~\cite{kingma2013auto} & 0.744 & - & 0.790 & - \\
    BSA~\cite{noguchi2019image} & 0.345 & 0.785 & 0.342 & 0.908   \\ \midrule
    MineGAN (w/o FT) & 0.349 & 0.774  & 0.347 & 0.891  \\
    MineGAN & \textbf{0.337} & \textbf{0.812} & \textbf{0.334} & \textbf{0.934}  \\ 
                \bottomrule
            \end{tabular}
            }
    \end{tabular} }
    \end{adjustbox}
    \vspace{-3mm}
    \caption{\small Results for various knowledge transfer methods. (Top) Generated images. (Bottom) KMMD and MV. \vspace{-3mm}}
	\label{fig:metrics_KMMD_SNGAN}
\end{figure}

We also compare here with Batch Statistics Adaptation (BSA)~\cite{noguchi2019image} using the same settings and architecture, namely SNGAN~\cite{miyato2018spectral}. 
They performed knowledge transfer from a pretrained SNGAN on ImageNet~\cite{krizhevsky2012imagenet} to FFHQ~\cite{karras2019style} and to Anime Face~\cite{danbooru2018}. 
Target domains have only 25 images of size 128$\times$128. 
We added our results to those reported in~\cite{noguchi2019image} in Fig.~\ref{fig:metrics_KMMD_SNGAN}~(bottom). 
Compared to BSA, MineGAN~(w/o FT) obtains similar KMMD scores, showing that generated images obtain comparable quality. 
MineGAN outperforms BSA both in KMMD score and Mean Variance. The qualitative results (shown in Fig.~\ref{fig:metrics_KMMD_SNGAN}~(top)) clearly show that MineGAN outperforms the baselines. 
BSA presents blur artifacts, which are probably caused by the mean square error used to optimize their model.

\begin{table}[t]
    \centering
          \resizebox{1\columnwidth}{!}{
            \centering
            \begin{tabular}{lccc}
                \toprule
                  Method & $\rightarrow$ Red vehicle & $\rightarrow$ Tower & $\rightarrow$ Bedroom \\ 
                  \midrule
                  Scratch & 190 / 185 / 196 & 176 & 181\\
                  TransferGAN (car) & 76.9 / 72.4 / 75.6 & - & - \\
                  TransferGAN (bus) & 72.8 / 71.3 / 73.5 & - & - \\
                  TransferGAN (livingroom) & - & 78.9 & 65.4 \\
                  TransferGAN (church) & - & 73.8 & 71.5 \\
                  MineGAN (w/o FT) & 67.3 / 65.9 / 65.8 & 69.2 & 58.9 \\
                  MineGAN & 61.2 / 59.4 / 61.5 &62.4 & 54.7\\
                  \midrule
                  \midrule
                  Estimated $p_i$ & & & \\
                  \midrule
                  Car & 0.34 / 0.48 / 0.64 & - & - \\ 
                  Bus & 0.66 / 0.52 / 0.36 & - & - \\
                  Living room & - & 0.07 & 0.45 \\
                  Kitchen & - & 0.06 & 0.40 \\
                  Bridge & - & 0.42 & 0.08 \\
                  Church & - & 0.45 & 0.07 \\
                \bottomrule  
            \end{tabular}
            }
            \caption{\small Results for $\{$Car, Bus$\}$ $\rightarrow $ Red vehicles with three different target data distributions (ratios cars:buses are 0.3:0.7, 0.5:0.5 and 0.7:0.3) and $\{$Living room, Bridge, Church, Kitchen$\}$ $\rightarrow $ Tower/Bedroom. 
    (Top) FID scores between real and generated samples. (Bottom) Estimated probabilities $p_i$ for each model. \vspace{-1mm}}  \label{table:lsun_fid_pro}
\end{table}

\begin{figure*}[t]
    \centering
    \includegraphics[width=\textwidth]{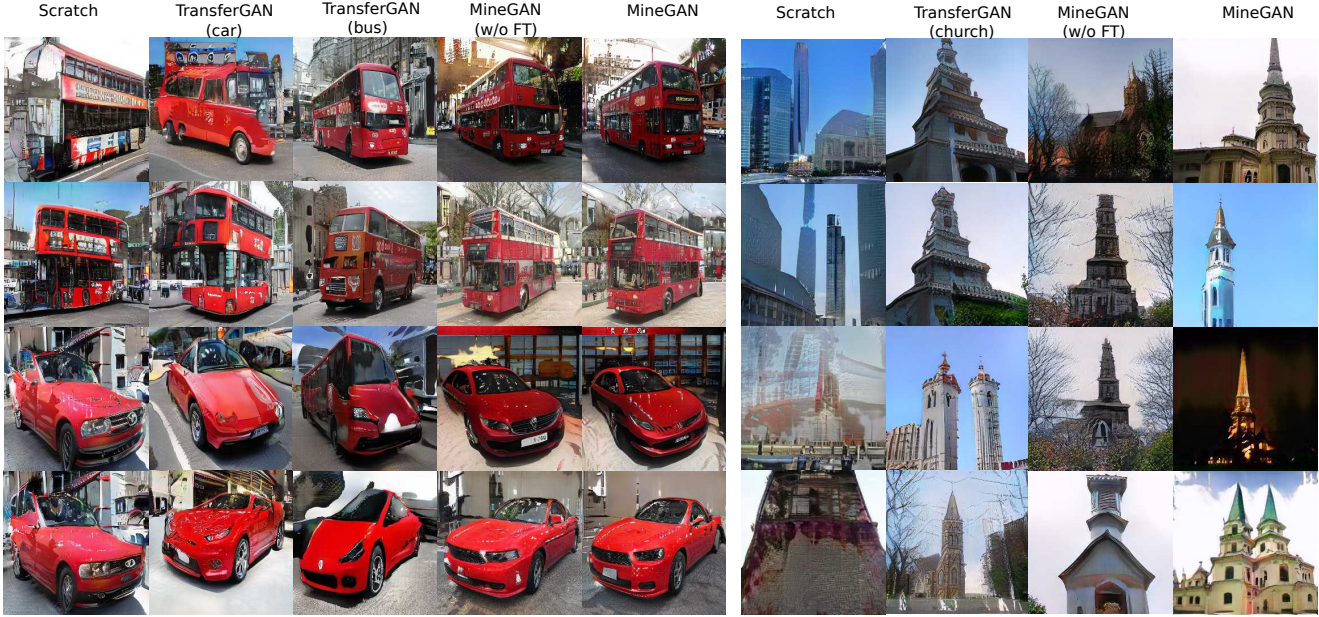}\vspace{-1mm}
    \caption{\small Results: $\{$car, bus$\}$ $\rightarrow $ red vehicles (left) and $\{$Living room, Bridge, Church, Kitchen$\}$ $\rightarrow $ Tower (right). Based on pretrained Progressive GAN. For TransferGAN we show the pretrained model between parentheses. More examples in Suppl, Mat~(Sec.~F). }\vspace{-5mm}
    \label{fig:target_vehicles}
    
\end{figure*}

\minisection{Multiple pretrained models.}
We now evaluate the general case for MineGAN, where there is more than one pretrained model to mine from.
We start with two pretrained Progressive GANs: one on \emph{Cars} and one on \emph{Buses}, both from the  LSUN dataset~\cite{yu2015lsun}. These pretrained networks generate cars and buses of a variety of different colors. We collect a target dataset of 200 images (of $256 \times 256$ resolution) of \emph{red vehicles}, which contains both red cars and red buses. 
We consider three target sets with different car-bus ratios (0.3:0.7, 0.5:0.5, and 0.7:0.3) which allows us to evaluate the estimated probabilities $p_i$ of the selector. 
To successfully generate \emph{all types} of red vehicle, knowledge needs to be transferred from both pre-trained models.

Fig.~\ref{fig:target_vehicles} shows the synthesized images. As expected, the limited amount of data makes training from scratch result in overfitting.   
TransferGAN~\cite{wang2018transferring} produces only high-quality output samples for one of the two classes (the class that coincides with the pretrained model) and it cannot extract knowledge from both pretrained GANs. On the other hand, MineGAN generates high-quality images by successfully transferring the knowledge from both source domains simultaneously. 
Table~\ref{table:lsun_fid_pro}~(top rows) quantitatively validates that our method outperfroms TransferGAN with a significantly lower FID score. 
Furthermore, the probability distribution predicted by the selector, reported in Table~\ref{table:lsun_fid_pro}~(bottom rows), matches the class distribution of the target data. 

To demonstrate the scalability of MineGAN with multiple pretrained models, we conduct experiments using four different generators, each trained on a different LSUN category including \textit{Livingroom}, \textit{Kitchen}, \textit{Church}, and \textit{Bridge}.
We consider two different off-manifold target datasets, one with \textit{Bedroom} images and one with \textit{Tower} images, both containing 200 images. Table~\ref{table:lsun_fid_pro}~(top) shows that our method obtains significantly better FID scores even when we choose the most relevant pretrained GAN to initialize training for TransferGAN. Table~\ref{table:lsun_fid_pro}~(bottom) shows that the miner identifies the relevant pretrained models, e.g. transferring knowledge from \textit{Bridge} and \textit{Church} for the target domain \textit{Tower}. Finally, Fig.~\ref{fig:target_vehicles}~(right) provides visual examples.

\begin{figure*}[t]
    \centering
    \includegraphics[width=\textwidth]{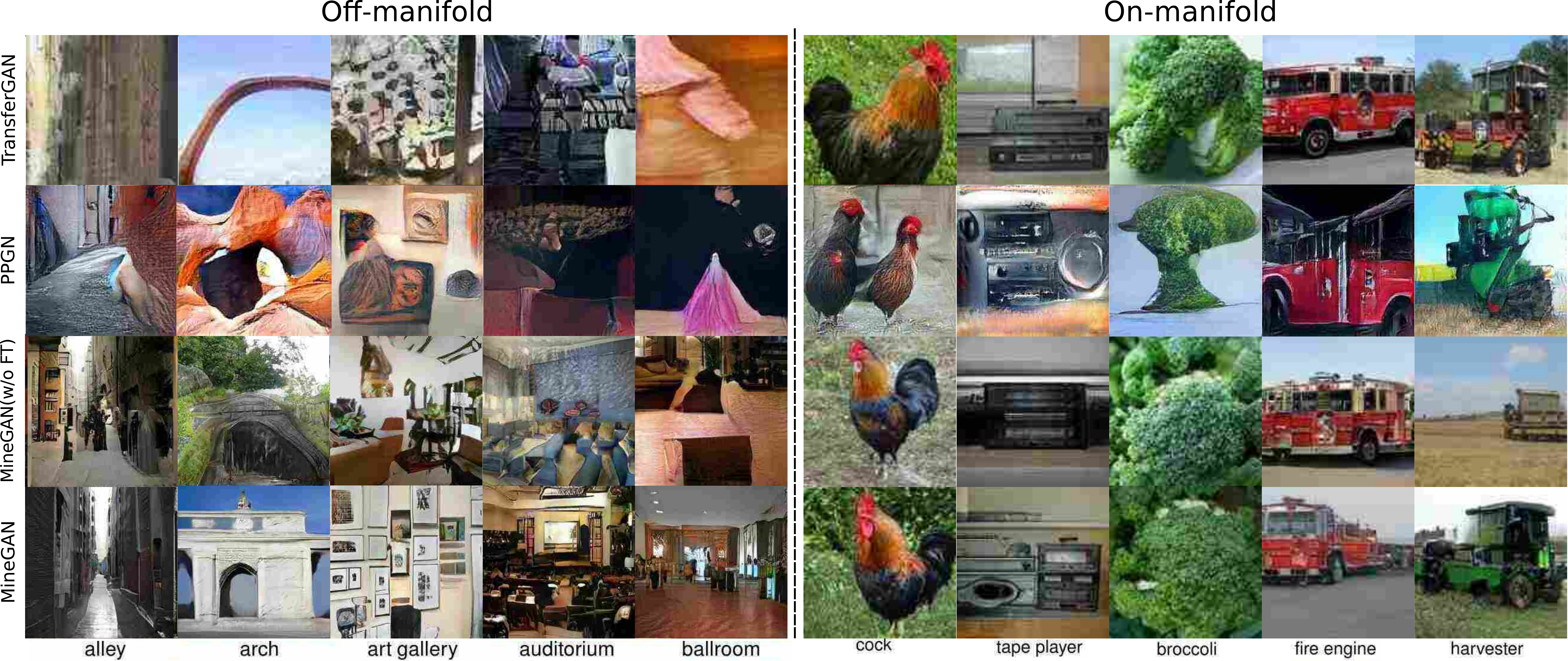}
    \caption{\small Results for conditional GAN. (Left) Off-manifold (ImageNet$\rightarrow$Places365). (Right) On-manifold (ImageNet$\rightarrow$ImageNet).} \vspace{-3mm}
    \label{fig:conditional_imagenet_places}
\end{figure*}

\begin{table}[t]
    \centering
   
    \begin{subtable}{1\linewidth}
      \centering
          \resizebox{1\columnwidth}{!}{
            \begin{tabular}{lccccc}
            \toprule
                            \multirow{2}{*}{Method}  & \multicolumn{2}{c}{ Off-manifold}  & \multicolumn{2}{c}{ On-manifold}& \multirow{2}{*}{Time (ms)}\\
                            & Label & FID/KMMD & Label & FID/KMMD & \\ 
                    \midrule
                 Scratch & No &190 / 0.96&No&187 / 0.93&5.1 \\
                 TransferGAN & No &89.2 / 0.53&Yes&58.4 / 0.39&5.1 \\
                 DGN-AM & Yes &214 / 0.98&Yes&180 / 0.95&3020\\ 
                 PPGN& Yes & 139 / 0.56&Yes&127 / 0.47&3830\\ 
                 MineGAN (w/o FT)& No& 82.3 / 0.47 &No&61.8 / 0.32&5.2 \\ 
                 MineGAN& No& 78.4 / 0.41 &No&52.3 / 0.25&5.2 \\ 
                \bottomrule
            \end{tabular}
            }

    \end{subtable} \vspace{-3mm}
    \caption{\small  Distance between real data and generated samples as measured by FID score and KMMD value. The off-manifold results correspond to ImageNet $\rightarrow $ Places365, and the on-manifold results correspond to ImageNet $\rightarrow $ ImageNet. We also indicate whether the method requires the target label. Finally, we show the inference time for the various methods in milliseconds. \vspace{-4mm}}
    \label{table:appendix_target_imagenet_places365_fid}
\end{table}

\subsection{Knowledge transfer from conditional GANs}
\label{sec:conditional}
\vspace{-2mm}

Here we transfer knowledge from a pretrained \emph{conditional GAN} (see Section~\ref{sec:multipleGANs}).
We use BigGAN~\cite{brock2018large}, which is trained using ImageNet~\cite{russakovsky2015imagenet}, and evaluate on two target datasets: on-manifold (ImageNet: \textit{cock}, \textit{tape player}, \textit{broccoli},  \textit{fire engine},  \textit{harvester}) and off-manifold (Places365~\cite{zhou2014object}:  \textit{alley},  \textit{arch},  \textit{art gallery},  \textit{auditorium}, \textit{ballroom}). 
We use 500 images per category. We compare MineGAN with training from scratch, TransferGAN~\cite{wang2018transferring}, and two iterative methods: DGN-AM~\cite{nguyen2016synthesizing} and PPGN~\cite{nguyen2017plug} \footnote{We were unable to obtain satisfactory results with BSA~\cite{noguchi2019image} in this setting (images suffered from  blur artifacts) and have excluded it here.}.
It should be noted that both DGN-AM~\cite{nguyen2016synthesizing} and  PPGN~\cite{nguyen2017plug} are based on a less complex GAN (equivalent to DCGAN~\cite{radford2015unsupervised}).
Therefore, we expect these methods to exhibit results of inferior quality, and so the comparison here should be interpreted in the context of GAN quality progress. However, we would like to stress that both DGN-AM and PPGN do not aim to transfer knowledge to new domains. 
They can only generate samples of a particular class of a pretrained classifier network, and they have no explicit loss ensuring that the generated images follow a target distribution. 

Fig.~\ref{fig:conditional_imagenet_places} shows qualitative results for the different methods.  
As in the unconditional case, MineGAN produces very realistic results, even for the challenging off-manifold case. Table~\ref{table:appendix_target_imagenet_places365_fid} presents quantitative results in terms of FID and KMMD.
We also indicate whether each method uses the label of the target domain class.
Our method obtains the best scores for both metrics, despite not using target label information.
PPGN performs significantly worse than our method.
TransferGAN has a large performance drop for the off-manifold case, for which it cannot use the target label as it is not in the pretrained GAN (see~\cite{wang2018transferring} for details).

Another important point regarding DGN-AM and PPGN is that each image generation during inference is an iterative process of successive backpropagation updates until convergence, whereas our method is feedforward.
For this reason, we include in Table~\ref{table:appendix_target_imagenet_places365_fid} the inference running time of each method, using the default 200 iterations for DGN-AM and PPGN. 
All timings have been computed with a CPU Intel Xeon E5-1620 v3 @ 3.50GHz and GPU NVIDIA RTX 2080 Ti.
We can clearly observe that the feedforward methods (TransferGAN and ours) are three orders of magnitude faster despite being applied on a more complex GAN~\cite{brock2018large}. 

\section{Conclusions}
We presented a model for knowledge transfer for generative models. It is based on 
a mining operation that identifies the regions on the learned GAN manifold that are closer to a given target domain. Mining leads to more effective and efficient fine tuning, even with few target domain images. Our method can be applied to single and multiple pretrained GANs. Experiments with various GAN architectures (BigGAN, Progressive GAN, and SNGAN) on multiple datasets demonstrated its effectiveness. 
Finally, we demonstrated that MineGAN can be used to transfer knowledge from multiple domains. 

\minisection{Acknowledgements.} We acknowledge the support from Huawei Kirin Solution, the Spanish projects TIN2016-79717-R and RTI2018-102285-A-I0, the CERCA Program of the Generalitat de Catalunya, and the EU Marie Sklodowska-Curie grant agreement No.6655919.

{\small
\bibliography{shortstrings,ref}
\bibliographystyle{ieee_fullname}
}
\appendix

\section{Architecture and training details}
\label{Appendix_training_details}

\vspace{-1mm}
\textbf{MNIST dataset.} Our model contains a \textit{miner}, a \textit{generator} and a \textit{discriminator}. For both unconditional and conditional GANs, we use the same framework~\cite{gulrajani2017improved} to design the generator and discriminator. The miner is composed of two fully connected layers with the same dimensionality as the latent space $|z|$. The visual results are computed with $|z|=16$; we found that the quantitative results improved for larger $|z|$ and choose $|z|=128$.

We randomly initialize the weights of each miner following a Gaussian distribution centered at 0 with 0.01 standard deviation, and optimize the model using Adam~\cite{kingma2014adam} with batch size of 64. The learning rate of our model is 0.0004,  with exponential decay rates of $\left ( \beta_{1},   \beta_{2} \right ) = \left ( 0.5, 0.999 \right )$. 

In the conditional MNIST case, label $c$ is a one-hot vector. 
This differs from the conditioning used for BigGAN~\cite{brock2018large} explained in Section~3.3.

Here we extend MineGAN to this type of conditional models by considering each possible conditioning as an independently pretrained generator and using the selector to predict the conditioning label. 
Given a conditional generator $G(c,z)$, we consider $G(i,z)$ as $G_i$ and apply the presented MineGAN approach for multiple pretrained generators on the family $\{G(i,z)| \text{ } i=1,...,N\}$.
The resulting selector now chooses among the $N$ classes of the model rather than among $N$ pretrained models, but the rest of the MineGAN training remains the same, including the training of $N$ independent miners.

\textbf{CelebA Women,  FFHQ Children and LSUN (Tower and Bedroom) datasets.} We design the generator and discriminator based on Progressive GANs~\cite{karras2017progressive}. Both networks use a multi-scale technique to generate high-resolution images. Note we use a \emph{simple} miner for tasks with dense and narrow source domains. The miner comprises out of four fully connected layers (8-64-128-256-512), each of which is followed by a \textit{ReLU} activation and \textit{pixel normalization}~\cite{karras2017progressive} except for last layer.  We use a Gaussian distribution centered at 0 with 0.01 standard deviation to initialize the miner, and optimize the model using Adam~\cite{kingma2014adam} with batch size of 4. The learning rate of our model is 0.0015, with exponential decay rates of $\left ( \beta_{1},   \beta_{2} \right ) = \left ( 0, 0.99 \right )$.

\textbf{FFHQ Face and Anime Face datasets.} We use the same network as~\cite{miyato2018spectral}, namely SNGAN. The miner consists of three fully connected layers (8-32-64-128).  We randomly initialize the weights following a Gaussian distribution centered at 0 with 0.01 standard deviation.
 For this additional set of experiments, we use Adam~\cite{kingma2014adam} with a batch size of 8, following a hyper parameter learning rate of 0.0002 and  exponential decay rate of  $\left ( \beta_{1},   \beta_{2} \right ) = \left ( 0, 0.9 \right )$. 

\textbf{Imagenet and Places365 datasets.}  We use the pretrained BigGAN~\cite{brock2018large}. We ignore the projection loss in the discriminator, since we do not have access to the label of the target data.   We employ a more \emph{powerful} miner in order to allocate more capacity to discover the regions related to target domain. The miner consists of two sub-networks:  miner $M^z$ and  miner $M^c$. Both $M^z$ and  $M^c$ are composed of four fully connected layers of sizes (128, 128)-(128, 128)-(128, 128)-(128, 128)-(128, 120) and (128, 128)-(128, 128)-(128, 128)-(128, 128)-(128, 128), respectively. 
We use Adam~\cite{kingma2014adam} with a batch size of 256, and learning rates of 0.0001 for the miner and the generator and 0.0004 for the discriminator. 
The exponential decay rates are $\left ( \beta_{1},   \beta_{2} \right ) = \left ( 0, 0.999 \right )$. We randomly initialize the weights following a Gaussian distribution centered at 0 with 0.01 standard deviation.

The input of BigGAN~\cite{brock2018large} is a random latent vector and a class label that is mapped to an embedding space. 
We therefore have two miner networks, the original one that maps to the input latent space ($M^z$) and a new one that maps to the latent class embedding ($M^c$). Note that since we have no class label, the miner ($M^c$) needs to learn what distribution over the embeddings best represents the target data.

\section{Evaluation metric details} 
\vspace{-2mm}
Similarly to~\cite{noguchi2019image}, we compute FID between 10,000 randomly generated images and 10,000 real images, if possible.
When the number of target image exceeds 10,000, we randomly select a subset containing only 10,000. 
On the other hand, if the target set contains fewer images, we cap the amount of randomly generated images to this number to compute the FID. Note we also consider KMMD to evaluate the distance between generated images and real images since FID suffers from instability on small datasets.

\section{Ablation study}
\label{sec:appendix_ablation}
In this section, we evaluate the effect of each independent contribution to MineGAN and their combinations. 

\minisection{Selection strategies.}
We ablate the use of the $\max$ operation in Eqs.~(5) and~(6), and replace it with a mean operation.
We call this setting MineGAN (mean). 
In this case the backpropagation is not only performed for the image with the highest critic score but for all images. We hypothesized that the $\max$ operation is necessary to correctly estimate the  probabilities used by the selector. We report the results in Table~\ref{table:appendix_lsun_fid_pro}, where we refer to our original model as MineGAN ($\max$). 
The probability distribution predicted by the selector indicates that MineGAN (mean) equally chooses both pretrained models on Car and Bus, while MineGAN successfully estimates the class distribution of the target data.

\minisection{Miner Architecture.}
\begin{figure}
    \centering
    \includegraphics[width=\columnwidth]{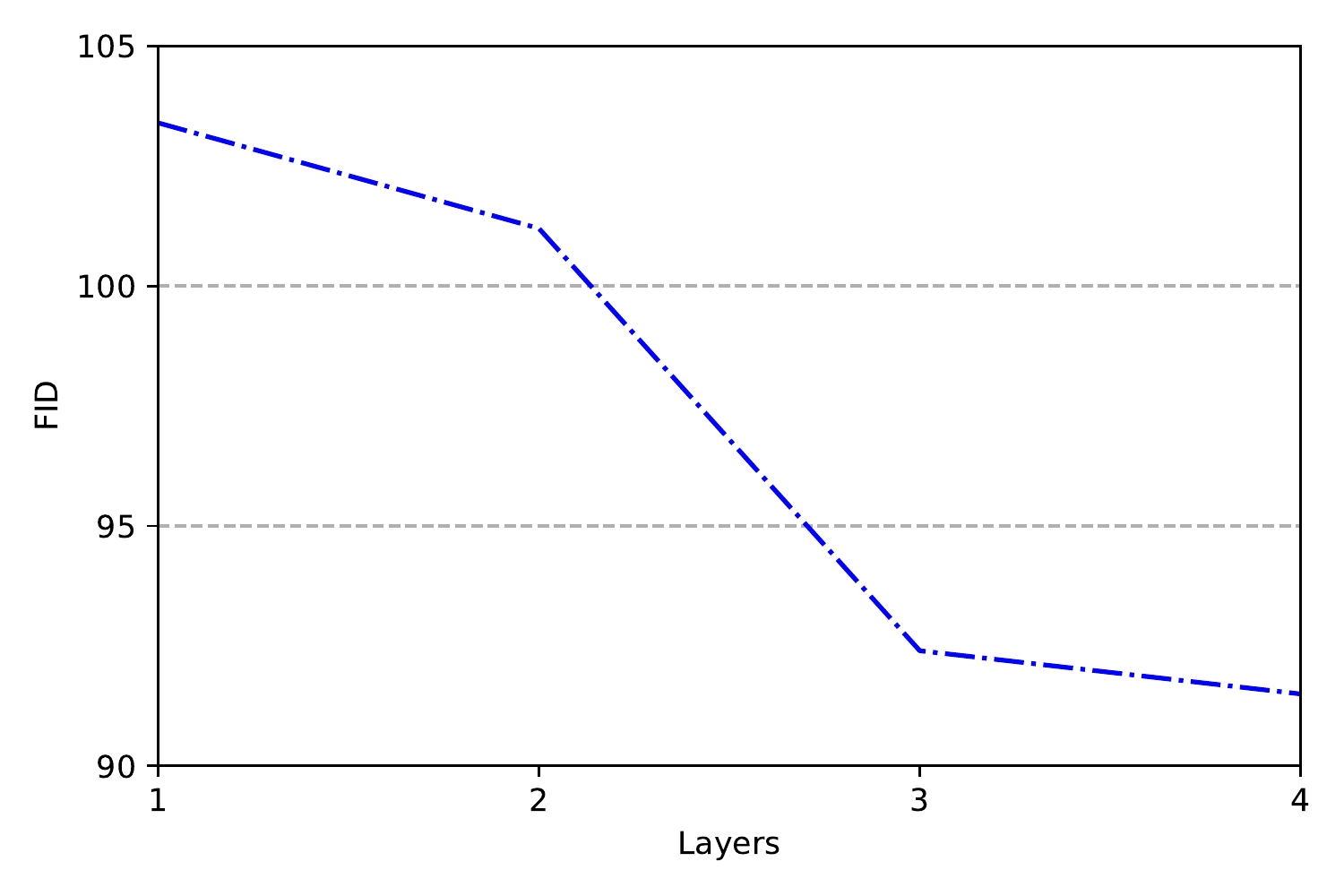}\vspace{-3mm}
    \caption{FID values for different number of fully connected layers in the miner. Results are based on the pretrained \emph{BigGAN} with target class \emph{Arch}.}
    \label{fig:appendix_FID_arch}
\end{figure}
We performed an ablation study for different variants of the miner based on pretrained BigGAN.
The miner always contains only fully connected layers, but we experiment with varying the number of layers. 
In Fig.~\ref{fig:appendix_FID_arch}, we show the results on off-manifold target class \textit{Arch} from Places365~\cite{zhou2014learning}. %
Using more layers increases the performance of the method. 
Besides, we find that the results of both 3 and 4 layers are similar, indicating that adding additional layers would only result in slight improvements for our model. 
Therefore, in this paper, we use miners with 4 fully connected layers.

\begin{table}[t]
    \centering
          \resizebox{1\columnwidth}{!}{
           \setlength{\tabcolsep}{20pt}
            \centering
            \begin{tabular}{lcc}
                \toprule
                  Method & MineGAN (mean)  & MineGAN ($\max$) \\ 
                  \midrule
                  Car & 0.51  & 0.34 \\ 
                  Bus & 0.49  & 0.66 \\
                \bottomrule  
            \end{tabular}
            }
            \caption{\small Estimated probabilities $p_i$ for $\{$Car, Bus$\}$ $\rightarrow $ Red vehicles for MineGAN with mean or $\max$ in Eqs.~(5) and~(6). The actual data distribution is 0.3:0.7 (ratio cars:buses). \vspace{-5mm}}  \label{table:appendix_lsun_fid_pro}
\end{table}

\section{MNIST experiment}
\label{sec:appendix_mnist}
We expand the MNIST experiments presented in Section~5.1 by providing a quantitative evaluation and including results on conditional GANs.
As evaluation measures, we use FID (Section~5) and classifier error~\cite{shmelkov2018good}. 
To compute classifier error, we first train a CNN classifier on real training data to distinguish between multiple classes (e.g.\ digit classifier).
Then, we classify the generated images that should belong to a particular class and measure the error as the percentage of misclassified images. 
This gives us an estimation of how realistic and accurate the generated images are in the context of targeted generation.

\begin{figure}[t]
    \centering
    \begin{tabular}{ c c c c c c} 
    \includegraphics[width=.15\textwidth,height=1.75cm]{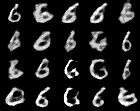}
    \includegraphics[width=.15\textwidth,height=1.75cm]{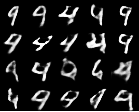}
    \includegraphics[width=.15\textwidth,height=1.75cm]{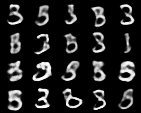}\\
    \includegraphics[width=.15\textwidth,height=1.75cm]{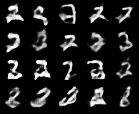}
    \includegraphics[width=.15\textwidth,height=1.75cm]{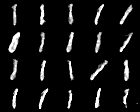}
    \includegraphics[width=.15\textwidth,height=1.75cm]{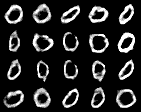}
    \end{tabular}
    \vspace{-3mm}
    \caption{Results for unconditional off-manifold generation of digits `6', `4', `3', `2', `1', `0'.}
    \label{fig:target_MNIST2}
\end{figure}

Table~\ref{tab:mining_results} presents the results for both unconditional and conditional models, using a noise length of $|z|=128$.
The relatively low error values indicate that the miner manages to identify the correct regions for generating the target digits.
The conditional model offers better results than the unconditional one by selecting the target class more often. 
We can also observe that the off-manifold task is more difficult than the on-manifold task, as indicated by the higher evaluation scores.
However, the off-manifold scores are still reasonably low, indicating that the miner manages to find suitable regions from other digits by mining local patterns shared with the target. 
Overall, these results indicate the effectiveness of mining on MNIST for both types of targeted image generation. In addition, in Fig.~\ref{fig:target_MNIST2} we have added a visualization for the off-manifold MNIST classes which were not already shown in Fig.~2. 

\begin{table}[t]
    \centering
    \resizebox{0.5\textwidth}{!}{
    \begin{tabular}{lcccc}
    \toprule
         \multirow{2}{*}{$d$} & \multicolumn{2}{c}{On-manifold} &  \multicolumn{2}{c}{Off-manifold} \\
        & Unconditional & Conditional & Unconditional & Conditional\\  
        \midrule
        0 & 13.4 / 2.5 & 12.6 / 0.7& 21.3 / 2.8 & 15.6 / 1.1\\
        1 & 13.1 / 1.7 & 12.6 / 1.9 & 15.9 / 2.5 & 14.8 / 2.1\\
        2 & 14.6 / 6.3 & 12.8 / 2.7 & 23.1 / 5.2 & 18.2 / 3.6\\
        3 & 14.1 / 10.1 & 13.3 / 1.6 & 22.8 / 7.3 & 14.2 / 1.5\\
        4 & 14.7 / 6.4 & 13.4 / 1.2 & 23.4 / 6.3 & 15.3 / 4.2\\
        5 & 13.1 / 9.3 & 11.7 / 2.1 & 21.9 / 10.9 & 17.2 / 5.7 \\
        6 & 13.4 / 2.8 & 14.3 / 1.8 & 24 / 3.1 & 15.8 / 1.6 \\
        7 & 12.9 / 3.2 & 14.2 / 1.8 & 24.8 / 4.9 & 16.3 / 2.6 \\
        8 & 14.2 / 7.5 & 14.7 / 5.5 & 25.7 / 9.8 & 18.7 / 5.6\\
        9 & 11.3 / 6.8 & 11.2 / 2.9 & 12.5 / 7.4 & 16.3 / 3.5 \\
        \midrule
        Average & 13.5 / 5.7 & 13.1 / 2.2 & 21.5 / 6.0 & 16.2 / 3.2 \\
    \bottomrule
    \end{tabular}}    
    \caption{\small Quantitative results of mining on MNIST, expressed as FID / classifier error.}
    \label{tab:mining_results}
\end{table}

\section{Further results on CelebA}
\label{Appendix_women_children}

\begin{figure*}[t]
    \centering
    \includegraphics[width=\textwidth,height=\textheight,keepaspectratio]{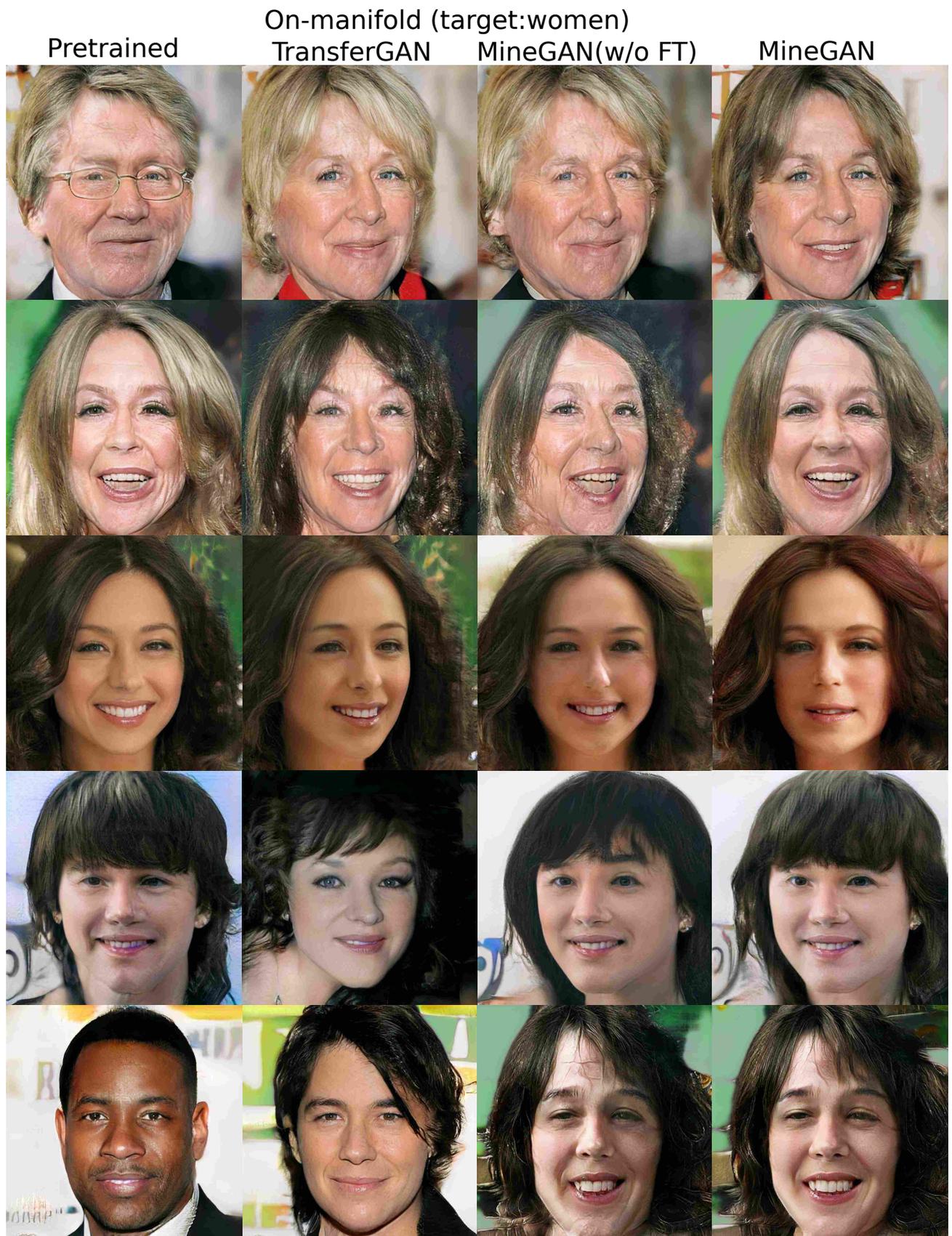}\vspace{-3mm}
    \caption{(CelebA$\rightarrow$FFHQ women). Based on pretrained \emph{Progressive GAN}.}
    \label{fig:appendix_target_hhfq_women}
\end{figure*}

\begin{figure*}[t]
    \centering
    \includegraphics[width=\textwidth,height=\textheight,keepaspectratio]{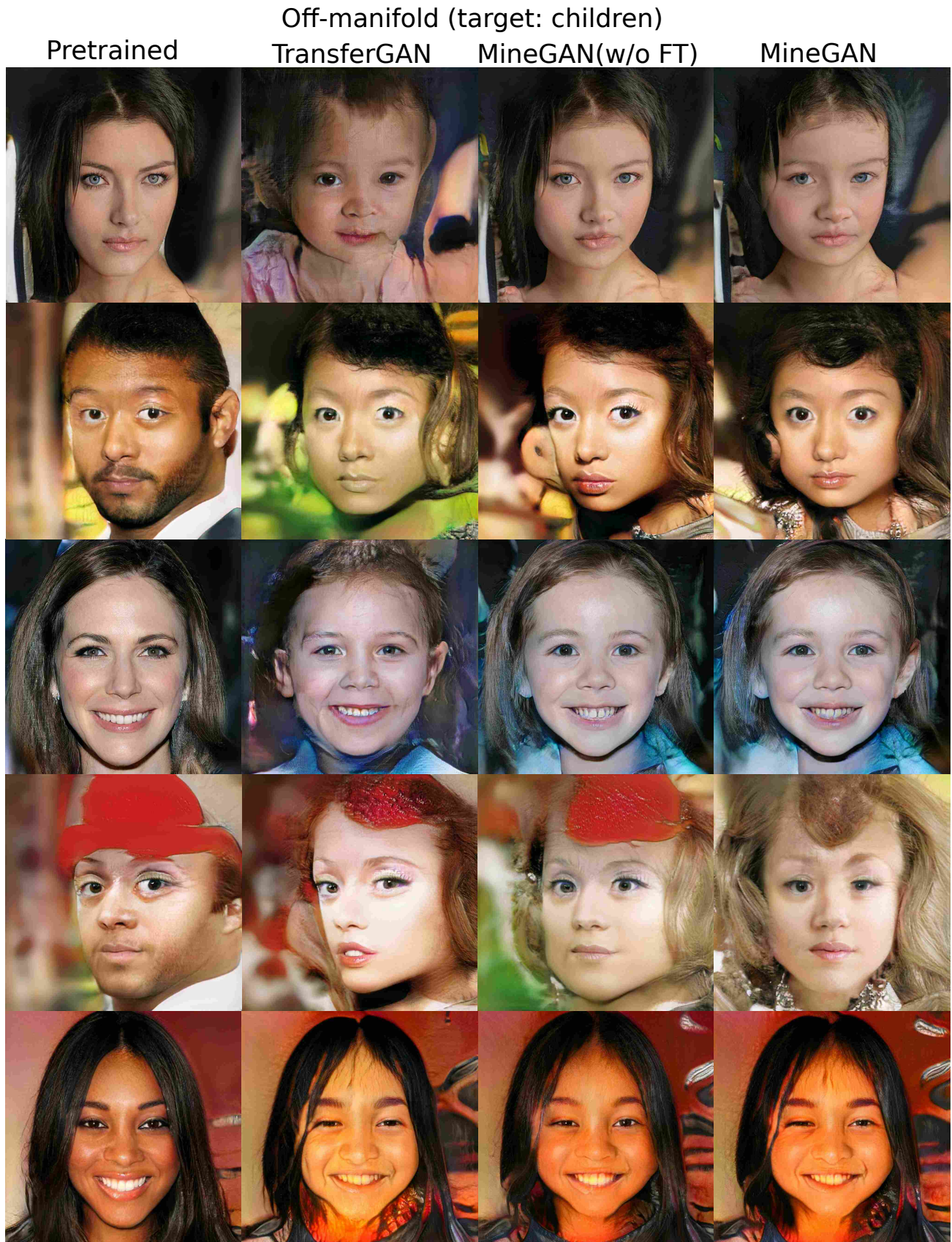}\vspace{-3mm}
    \caption{(CelebA$\rightarrow$ FFHQ children). Based on pretrained \emph{Progressive GAN}.}
    \label{fig:appendix_target_hhfq_children}
\end{figure*}

\begin{figure*}[t]
    \centering
    \includegraphics[width=\textwidth,height=\textheight,keepaspectratio]{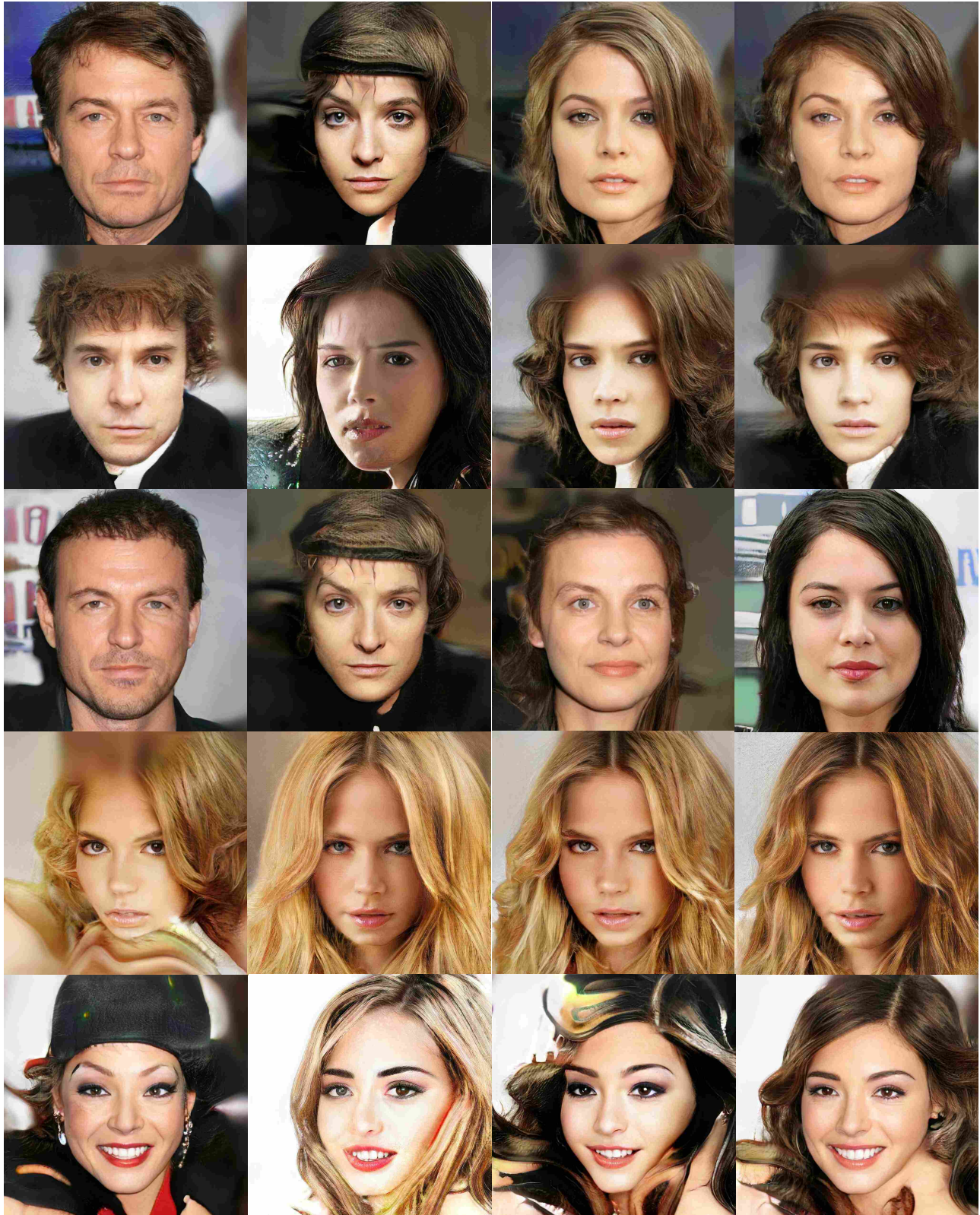}\vspace{-3mm}
    \caption{(CelebA$\rightarrow$CelebA women). Based on pretrained \emph{Progressive GAN}.}
    \label{fig:appendix_target_celeba_women}
\end{figure*}

We provide additional results for the on-manifold experiment  CelebA$\rightarrow$FFHQ women in Fig.~\ref{fig:appendix_target_hhfq_women}, and the off-manifold CelebA$\rightarrow$FFHQ children in Fig.~\ref{fig:appendix_target_hhfq_children}. In addition, we have also performed an on-manifold experiment with CelebA$\rightarrow$CelebA  women, whose results are provided in Fig.~\ref{fig:appendix_target_celeba_women}.

\begin{figure*}[t]
    \centering
    \includegraphics[width=\textwidth,height=.97\textheight,keepaspectratio]{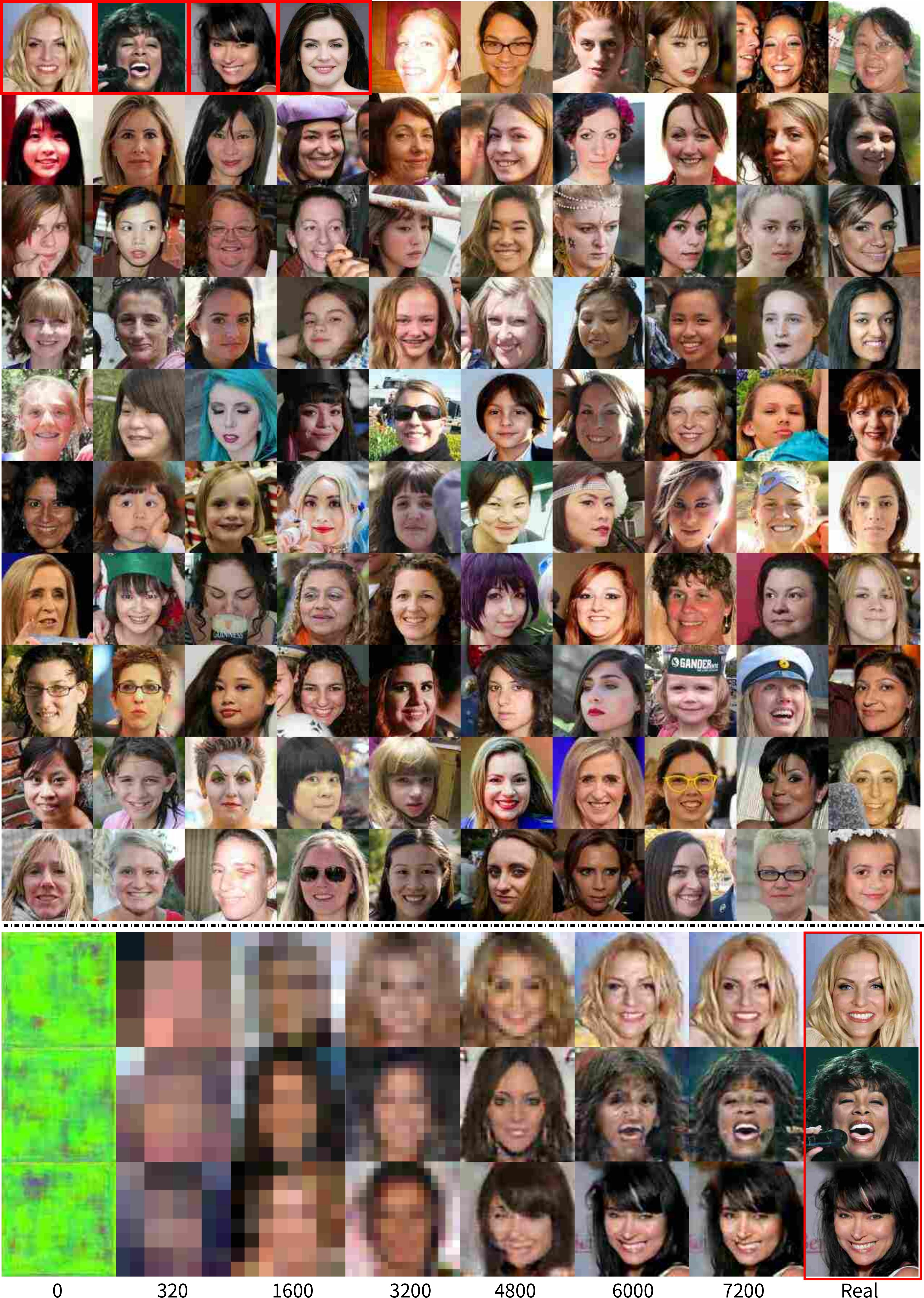}\vspace{-3mm}
    \caption{(Top) 100 women faces from HHFQ dataset. (Bottom) training of model from scratch: the images start with low quality and iteratively overfit to a particular training image. Red boxes identify images which are remembered by the model trained from scratch or from TransferGAN (see Fig. 4). Based on pretrained \emph{Progressive GAN}.}
    \label{fig:appendix_dataset_women}
\end{figure*}

\begin{figure*}[t]
    \centering
    \includegraphics[width=\textwidth,height=\textheight,keepaspectratio]{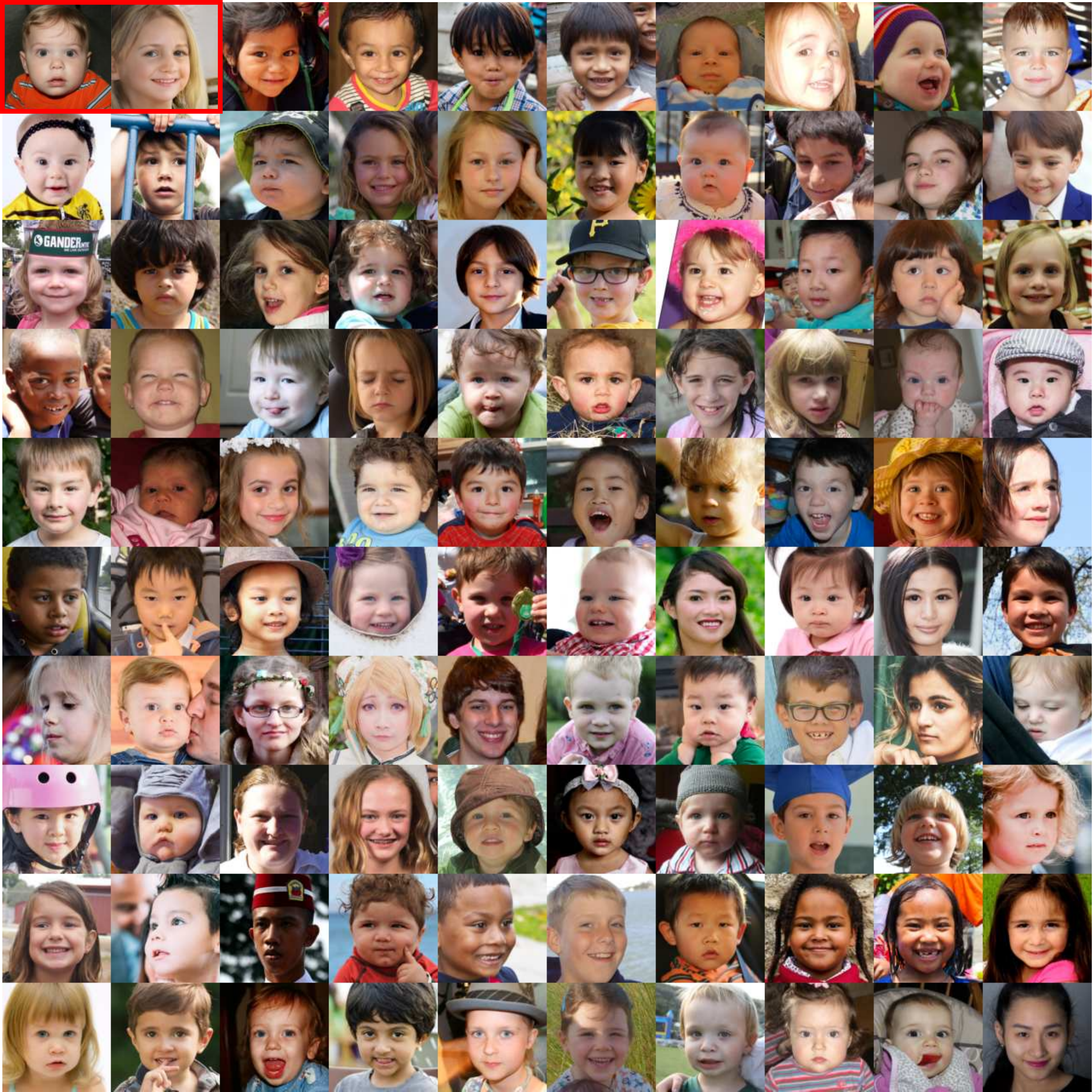}\vspace{-3mm}
    \caption{100 children faces from HHFQ dataset. Red boxes identify images which are remembered by the model trained from scratch  (see Fig. 4).  Based on pretrained \emph{Progressive GAN}.}
    \label{fig:appendix_dataset_women}
\end{figure*}

\vspace{-2mm}
\section{Further results for LSUN}
\label{Appendix_tower_bedroom}

\begin{figure*}[t]
    \centering
    \includegraphics[width=\textwidth,height=\textheight]{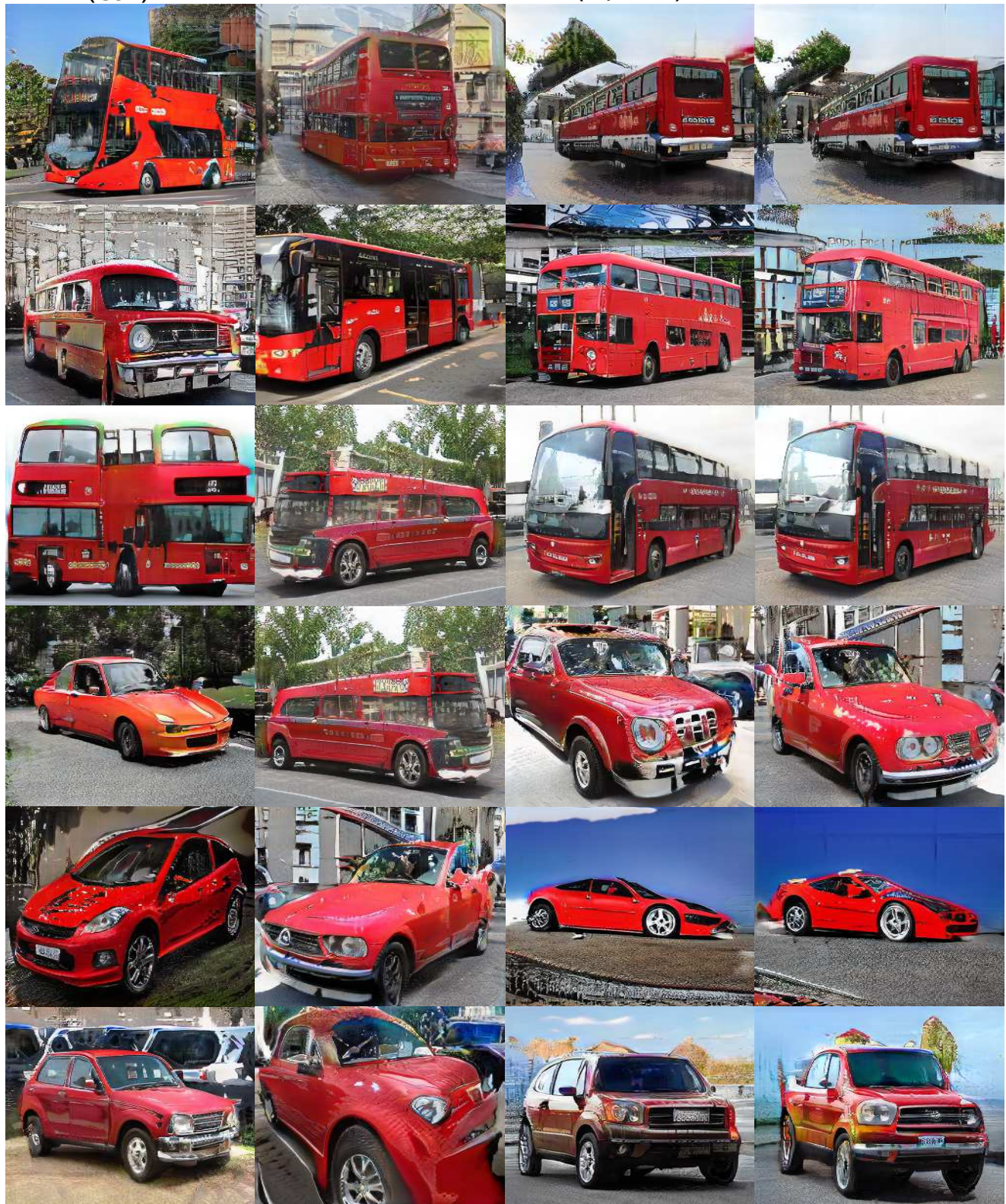}\vspace{-3mm} 
    \caption{($\{$bus, car$\}$) $\rightarrow $red vehicles. Based on pretrained \emph{Progressive GAN}.}\label{fig:appendix_target_vehicles}

\end{figure*}

\begin{figure*}[t]
    \centering
    \includegraphics[width=1\textwidth]{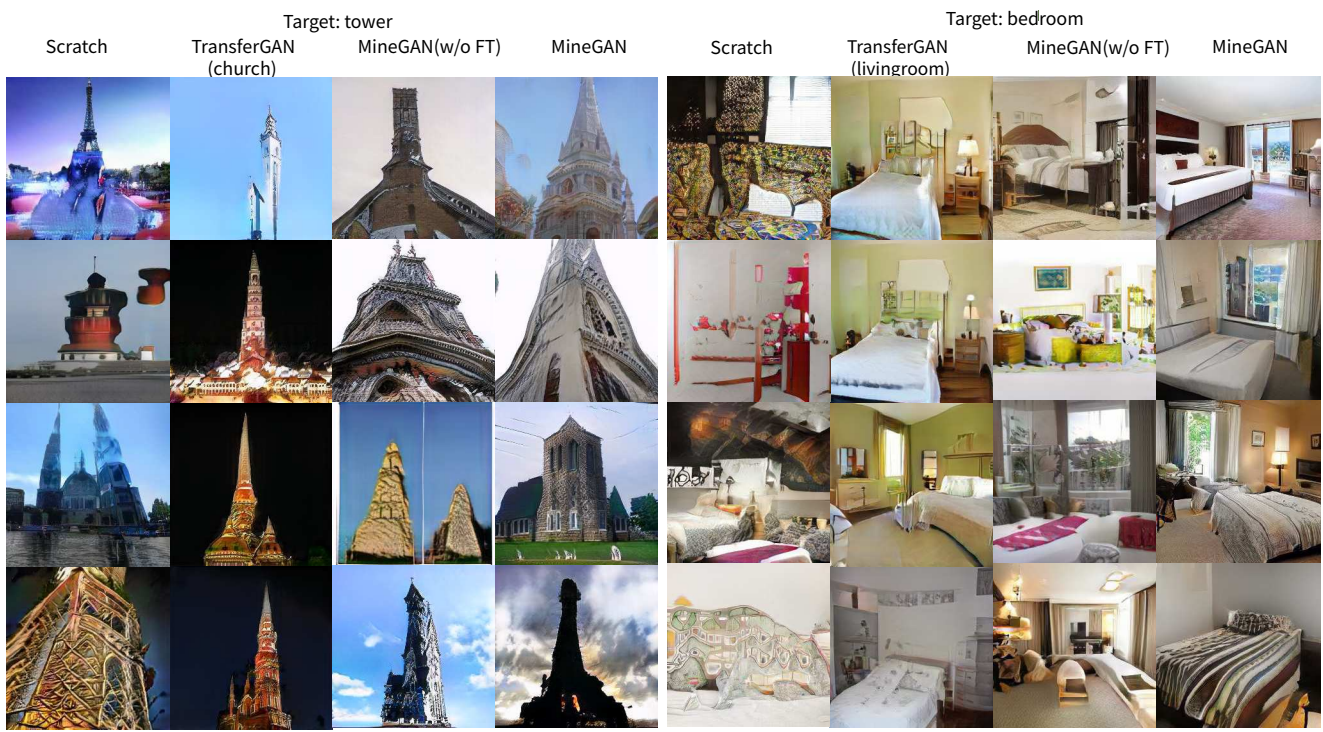}
    \caption{\small Results for unconditional GAN. (Top) (Livingroom, kitchen, bridge, church )$\rightarrow$Tower. (Bottom) (Livingroom, kitchen, bridge, church )$\rightarrow$Bedroom. Based on pretrained \emph{Progressive GAN}.}\label{fig:appendix_target_tower_bedroom}
\end{figure*}

We provide additional results for the experiment ($\{$Bus, Car$\}$) $\rightarrow $ Red vehicles in Fig.~\ref{fig:appendix_target_vehicles} and for the experiment  $\{$Bedroom, Bridge, Church, Kitchen$\}$ $\rightarrow $ Tower/Bedroom in Fig.~\ref{fig:appendix_target_tower_bedroom}.
When applying MineGAN to multiple pretrained GANs, we use one of the domains to initialize the weights of the critic. In Fig.~\ref{fig:appendix_target_tower_bedroom} we used 
\textit{Church} to initialize the critic in case of the target set \textit{Tower}, and \textit{Kitchen} to initialize the critic for the target set \textit{Bedroom}. We found this choice to be of little influence on the final results. When using \textit{Kitchen} to initialize the critic for target set \textit{Tower} results change from 62.4 to 61.7. When using \textit{Church} to initialize the critic for target set \textit{Bedroom} results change from 54.7 to 54.3.

\end{document}